%% file: chapgpt_fie_2023.tex
\documentclass[conference]{IEEEtran}
\IEEEoverridecommandlockouts
\usepackage{cite}
\usepackage{amsmath,amssymb,amsfonts}
\usepackage{algorithmic}
\usepackage{graphicx}
\usepackage{textcomp}
\usepackage{xcolor}
\usepackage{dirtytalk}
\usepackage{wasysym}
\usepackage{url}
\usepackage{multicol}
\usepackage{multirow}
\usepackage{booktabs}
\usepackage{makecell}
\usepackage{comment}

\newenvironment{myquote}%
  {\list{}{\leftmargin=0.25in\rightmargin=0.25in
  }\item[]}%
  {\endlist}

\begin{document}

\begin{minipage}{\textwidth}
\vspace{5em}
{\large
\textbf{IEEE Copyright Notice}\\[1em]
© 2024 IEEE.  Personal use of this material is permitted.  Permission from IEEE must be obtained for all other uses, in any current or future media, including reprinting/republishing this material for advertising or promotional purposes, creating new collective works, for resale or redistribution to servers or lists, or reuse of any copyrighted component of this work in other works.\\[1em]
Accepted to be published in: Proceedings of the 2024 IEEE ASEE Frontiers in Education Conference, Washington, D.C., October 13–16, 2024.
}
\end{minipage}
\newpage

\title{How Novice Programmers Use and Experience ChatGPT when Solving Programming Exercises in an Introductory Course}

\author{
\IEEEauthorblockN{Andreas Scholl}
\IEEEauthorblockA{\textit{Computer Science} \\
\textit{Nuremberg Tech}\\
Nuremberg, Germany \\
andreas.scholl@th-nuernberg.de}
\and
\IEEEauthorblockN{Natalie Kiesler}
\IEEEauthorblockA{\textit{Computer Science} \\
\textit{Nuremberg Tech}\\
Nuremberg, Germany \\
natalie.kiesler@th-nuernberg.de}
}

\maketitle

\begin{abstract} 
This research paper contributes to the computing education research community's understanding of Generative AI (GenAI) in the context of introductory programming, and specifically, how students utilize related tools, such as ChatGPT. An increased understanding of students' use is mandatory for educators and higher education institutions, as GenAI is here to stay, and its performance is likely to improve rapidly in the near future. Learning about students' use patterns is not only crucial to support their learning, but to develop adequate forms of instruction and assessment. With the rapid advancement of AI, its broad availability, and ubiquitous presence in educational environments, elaborating how AI can enhance learning experiences, especially in courses such as introductory programming is important. To date, most studies have focused on the educator's perspective on GenAI, its performance, characteristics, and limitations. However, the student perspective, and how they actually use GenAI tools in course contexts, has not been subject to a great number of studies. Therefore, this study is guided by the following research questions: (1) What do students report on their use pattern of ChatGPT in the context of introductory programming exercises? and (2) How do students perceive ChatGPT in the context of introductory programming exercises? To address these questions, computing students at a large German university were asked to solve programming tasks with the assistance of ChatGPT as part of their introductory programming course. Students (n=298) provided information regarding the use of ChatGPT, and their evaluation of the tool via an online survey. This research provides a comprehensive evaluation of ChatGPT-3.5's application by novice programmers in a higher education context. The findings reveal that while students widely adopt GenAI, their use varies significantly, ranging from acceptance of generated solutions to dynamic, and critical engagement. Therefore, this work has implications for educators designing guardrails or forms of instructions on the use of GenAI tools in the classroom. 
\end{abstract}


\section{Introduction}
While educators and researchers are still trying to grasp the full impact of Generative Artificial Intelligence (GenAI) and related tools on our educational system(s), students are quickly moving forward, and open to experimenting with Copilot, and ChatGPT~\cite{amani2023generative,raman2023university,prather2023wgfullreport}. Even though it is challenging to predict which programming competencies~\cite{kiesler2024modeling} students will require in the future and how to adapt curricula respectively~\cite{becker2023generative,kiesler2023beyond}, computing educators need to gain an understanding of how students currently apply GenAI tools. 

There is no doubt that GenAI tools, such as ChatGPT, are powerful and performing well in the context of computing, and particularly in introductory programming courses. An increasing body of research has been dedicated to showing its potential in solving CS1 and CS2 assignments~\cite{geng2023chatgpt,kiesler2023large,savelka2023large}. Another use case for students is the enhancement of programming error messages~\cite{Leinonen2023,Sarsa2022,macneil2022experiences,leinonen2023comparing}, and the generation of formative feedback~\cite{Bengtsson_Kaliff_2023,kiesler2023exploring,LMU-TEL/ADS2023,azaiz2024feedbackgeneration,roest2023nextstep}. Generating personalized feedback can potentially benefit both students, and educators, especially in large scale classes, such as introductory programming~\cite{jeuring2022towards,kiesler2023investigating}. At the same time, GenAI tools still produce misleading information for novice learners~\cite{kiesler2023exploring}.

At this point, however, hypothesizing about students' use cases is a debate mostly led by educators~\cite{alhossami2024socratic,joshi2024chatgpt,prather2024howinstructors}, as they are reflecting and discussing the alignment of their learning objectives, instruction, learning activities and assessments in the GenAI era. So far, few studies have investigated GenAI tools from the student perspective (e.g., trust or attitude towards LLMs~\cite{amoozadeh2024trust,rogers2024attitudes}), or how students use such tools in the context of a computing course~\cite{liu2024teaching,grande2024studentperspective}.
Due to the broad availability of GenAI tools, and their well-known limitations, it is crucial to investigate how and for which purposes computing students use such tools, and how they perceive the benefits and limitations of GenAI. 

It is thus \textbf{the goal} of this research to gather the student perspective on using GenAI tools, such as ChatGPT. This perspective encompasses use pattern, but also students' reflections upon their experiences with ChatGPT in the context of a curricular course. To gather the student perspective, we developed an exercise sheet as part of an introductory programming course at a large, public university (Goethe University Frankfurt, Germany) in the winter term 2023/24. Students were asked to solve programming tasks with the assistance of ChatGPT-3.5 (freely available in Dec. 2023), and to participate in an online survey to report their experiences.

The quantitative and qualitative analysis of the survey responses (n=298) provides novel insights into students' use of ChatGPT in a curricular course setting. The \textbf{contribution of this paper} is a summary of students' use patterns, and reflections on applying ChatGPT while solving programming exercises. The presented research thereby supports educator's understanding of students' applying GenAI tools. This is crucial to help inform and develop adequate forms of instruction and pedagogical approaches supporting the critical and reflective use of GenAI tools within computing education.

\section{Related Work}
\label{sec:related_work}
GenAI and related tools have taken the world by storm~\cite{prather2023wgfullreport}.
Due to the broad availability of GenAI tools, starting with the launch of OpenAI's ChatGPT in late November 2022, computing educators are discussing implications for higher education institutions, curricula, learning objectives, assignments, and assessments ever since~\cite{macneil2024discussing,becker2023generative,prather2023wgfullreport}. In this section, we summarize benefits and limitations of GenAI tools in introductory programming. Moreover, we present recent findings on computing students' perspective on GenAI tools.

\subsection{Potential Benefits and Limitations of GenAI in Introductory Programming Contexts}

The performance of GenAI tools can be impressive, especially in introductory programming exercises and respective exams~\cite{finnieansley22,finnieansley23}. Several studies showed the capabilities of tools based on Large Language Models (LLMs), such as Copilot and ChatGPT~\cite{geng2023chatgpt,kiesler2023large,savelka2023large}. For example, Geng et al.~\cite{geng2023chatgpt} treated ChatGPT as a student participating in an introductory-level functional language programming course. The tool achieved a grade B and was ranked as 155 out of 314 students. Kiesler and Schiffner~\cite{kiesler2023large} had ChatGPT 3.5 and GPT-4 generate solutions to 72 standard introductory programming tasks available on CodingBat~\cite{parlante2024}. ChatGPT-3.5 immediately solved the task correctly in 69 cases (95,8\%), GPT-4 in 68 tasks (94,4\%)~\cite{kiesler2023large}. Savelka et al.~\cite{savelka2023large} showed that GPT models are more successful when prompted in natural language, or when asked to fill in blanks. Multiple-choice questions requiring the analysis or reasoning about code were more challenging for the models~\cite{savelka2023large}.

Despite these promising results, many studies have reported limitations of GenAI tools, such as (re-)producing biases or inaccurate information~\cite{gill2024transformative}. Another obvious problem is its knowledge cutoff so the tools lack recent data, e.g., on the latest Python version, which may cause irritations. Hallucinations within the generated data, and the fact that it can bypass plagiarism detection tools are additional issues for students and educators. As GenAI tools challenge the integrity of academic work, educational institutions and faculty need to take action on several levels~\cite{prather2023wgfullreport,zhai2022chatgpt}.
Prather et al.~\cite{prather2023wgfullreport} emphasize the need for ongoing research to keep up with the fast pace of technological developments.

In the context of introductory programming education, GenAI tools have been subject to research w.r.t. their application by educators and students~\cite{Sarsa2022}, with a focus on students' potential use~\cite{macneil2022experiences,zastudil2023studentpersp,Leinonen2023,phung2023generating,zhang2022repairing,kiesler2023exploring,azaiz2024feedbackgeneration}. One of these use cases is the generation of code explanations in a web software development e-book~\cite{macneil2022experiences}. MacNeil et al.~\cite{macneil2022experiences} analyzed students' perceptions towards automatically generated line-by-line code explanations by Codex and GPT-3. The majority of students evaluated as the explanations as helpful~\cite{macneil2022experiences}. Leinonen et al.~\cite{Leinonen2023} found that programming error messages (i.e., by the compiler and interpreter) can be enhanced by GenAI, as they can help create novice-friendly, actionable explanations.

GenAI tools were also shown to analyze students' code, and fix errors~\cite{phung2023generating,zhang2022repairing}. Phung et al. \cite{phung2023generating} studied the application of LLMs with the goal of fixing syntax errors in Python. They developed a technique to receive high precision feedback~\cite{phung2023generating}.
In other studies, it was shown that GenAI tools are capable of generating individual, elaborate feedback~\cite{narciss2006} to students' solutions to programming exercises~\cite{kiesler2023exploring,LMU-TEL/ADS2023,azaiz2024feedbackgeneration}. 

For example, Kiesler et al.~\cite{kiesler2023exploring} qualitatively analyzed the feedback generated by ChatGPT-3.5 in response to authentic student solutions to introductory programming exercises. In their study, they identified stylistic elements, textual explanations of the cause of errors and their fix, illustrating examples, meta-cognitive and motivational elements. However, ChatGPT generated misleading information, claimed uncertainty in its responses, and requested more information, thereby indicating its limitations, especially for novice learners~\cite{kiesler2023exploring}. 

Similarly, Azaiz et al.~\cite{LMU-TEL/ADS2023} analyzed the output generated by ChatGPT in response to introductory programming exercises and student solutions. They summarize a number of challenges, e.g., w.r.t. the formatting of the output, recognizing correct solutions, and hallucinations of errors in the students' code. In a follow-up study with GPT-4 Turbo, Azaiz et al.~\cite{azaiz2024feedbackgeneration} conclude notable improvements of the generated feedback, as the outputs were more structured, consistent, and always personalized. Due to this potential, we currently see the development of new educational tools and environments incorporating GenAI tools to create novice-friendly code explanations~\cite{taylor2024dcchelperrorexplanations}, or to not give the solution away~\cite{kazemitabaar2024codeaid}.

\subsection{Computing Students' Perspective on GenAI tools}

All of the aforementioned studies focused on students' potential use of GenAI tools, and thus hypothetical application scenarios. Some of them used authentic student data (e.g., students' solutions) as an input to GenAI tools~\cite{azaiz2024feedbackgeneration,kiesler2023exploring}. However, few studies directly involved students, their perspective, or their actual use of GenAI. 

In a study by Prather et al.~\cite{prather2023s}, students' use of GitHub Copilot as part of an introductory programming assignment was evaluated through observations and interviews. They asked students about their perceptions of the tool's usability, challenges and benefits, and implications for the tool's design. Vaithilingam et al.~\cite{vaithilingam2022expectation} also explored the usability of Copilot via user study with 24 participants. The participants appreciated the ability of Copilot to provide a useful starting point, and help find information (instead of having to search for it online). At the same time, programmers reported challenges related to understanding, editing, and debugging code snippets generated by Copilot. Overall, they conclude that Copilot's design should be improved~\cite{vaithilingam2022expectation}.

The student perspective on program synthesizers including Copilot was gathered by Jayagopal et al.~\cite{jayagopal2022exploring}, who focused on its learnability. They observed and interviewed 22 novice programmers who worked with the respective systems, and evaluated the systems' design. Students, for example, reported that prompting is perceived as less confusing and more exciting than writing format specifications~\cite{jayagopal2022exploring}. 

Denny et al.~\cite{denny2023promptly} also investigated first year students' interactions (n=54) with the tool \textit{Promptly}. Promptly was designed to teach students how to prompt GenAI tools based on predefined problems. It executes and tests the code generated by an LLM. In the respective pilot study, Denny et al.~\cite{denny2023promptly} quantified students' interactions with ChatGPT, analyzed their interactions, and their reflections on the use of GenAI. However, only 15 chat protocols were analyzed. Moreover, the character of the pilot was exploratory with the goal of evaluating the relevance of prompting skills for students~\cite{denny2023promptly}. The focus was thus not on the identification of interaction pattern or use cases.

An observation of 20 experienced programmers and an analysis of their interactions with Copilot led to the distinction of two operation modes: acceleration and exploration mode~\cite{barke2023grounded}. Copilot is thus either used to achieve a goal faster, or to explore different solution options. Barke et al.~\cite{barke2023grounded} further conclude that over-reliance on Copilot can hinder task completion, and that multiple suggestions can lead to a high cognitive load. 

In a recent controlled experiment, ChatGPT's application in a CS1 classroom was investigated via the evaluation of 56 students' task performance, interactions with, and reflection upon using ChatGPT~\cite{xue2024does}. The study did not find evidence for ChatGPT's impact on students' performance, but students who used ChatGPT tended to solely rely on it (instead of exploring other resources). Students further seemed neutral towards ChatGPT, but they noted potential ethical issues, and varying performance depending on the tasks~\cite{xue2024does}. 

Students' (and instructors') perceptions towards GenAI in academia were also surveyed by Amani et al.~\cite{amani2023generative}. The student survey focused on the current use of GenAI tools and their general perceptions. 813 students at Texas A\&M university responded to the survey. As a result, students seem to value the high availability of the tools, while recognizing their potential to be misused and that academia must adapt. Forman et al. conducted a similar online survey exploring high school students' perceptions towards ChatGPT~\cite{Forman_Udvaros_Avornicului_2023}. The analysis of 71 responses revealed students' positive attitude towards GenAI technology, as they are assuming it to play a role long term. Moreover, students reported to rely on GenAI tools to help save time during assignments and projects. 
Raman et al.~\cite{raman2023university} investigated university students' adoption of ChatGPT by addressing the attributes relative advantage, compatibility, ease of use, observability, and trialability. The analysis of 288 student responses revealed that all five factors influence students' intentions to use ChatGPT. 
The last example of a student survey is the study conducted by Prather et al.~\cite{prather2023wgfullreport} in 2023. They gathered 171 computing student responses from 17 countries, about half of them being in their first year of studies. Most students reported using GenAI for text or code generation, paraphrasing or summarizing text, and debugging. Some students, however, refuse to use GenAI tools due to their risks and ethical concerns. Students' perceptions regarding the relevance of GenAI for their future careers were mixed~\cite{prather2023wgfullreport}.

\subsection{Research Gap}
Despite the rapid increase of studies focusing on the application of GenAI in computing education and, specifically, by novice learners of programming~\cite{leinonen2023comparing,macneil2022experiences,kiesler2023exploring,azaiz2024feedbackgeneration}, many of these studies are controlled, or define certain scenarios. 
They thus represent the educator's perspective or focus on usability and design aspects, and are not necessarily authentic for the context of introductory programming exercises and courses. 
At the same time, the studies surveying students broadly address university and high school students' perception towards GenAI tools~\cite{amani2023generative,raman2023university,Forman_Udvaros_Avornicului_2023,prather2023wgfullreport} but lack specific educational contexts.
The student perspective on using GenAI for programming exercises while enrolled in a curricular programming course at university level is thus not yet investigated and well understood.

\section{Methodology}
In this section, we outline the research questions and goals guiding the present study. In addition, we introduce the survey design, the context of the data collection within an introductory programming course, and present the data analysis method. 

\subsection{Research Questions and Goals}

The present study has the goal to gather the perspective of novice programmers using ChatGPT in the context of introductory programming exercises. Specifically, we are addressing the following research questions (RQs):
\begin{itemize}
    \item[1.] \textit{What do students report on their use patterns of ChatGPT in the context of introductory programming exercises?}
    \item[2.] \textit{How do students perceive ChatGPT in the context of introductory programming exercises?}
\end{itemize}
To answer these questions, the study leverages empirical data collected in a survey with computing students (n=298) enrolled in an introductory programming course at a large German university. Before completing the survey, students were asked to complete a set of programming exercises with the assistance of ChatGPT-3.5. 

\subsection{Survey Development}

The survey design was guided by the RQs and informed as far as possible by prior work. Most importantly, we wanted to gather students' use patterns when working on the exercises as part of the course (RQ1). Therefore, we asked students how often, how long and for which purposes they use ChatGPT in the problem solving process. For the latter we used some of the application scenarios identified in related work (e.g.,~\cite{phung2023generating,Leinonen2023,macneil2022experiences,LMU-TEL/ADS2023,vaithilingam2022expectation,prather2023wgfullreport}).

Similarly, the questions focusing on students' perceptions of ChatGPT (RQ2) were informed by prior work, for example, regarding ChatGPT's usability~\cite{denny2023promptly,jayagopal2022exploring,vaithilingam2022expectation,prather2023s}. Therefore, our survey addresses the concepts \textit{ease of use}, \textit{perceived skills gain}, \textit{accuracy and relevance of the responses}, and \textit{user satisfaction}. The survey also incorporates open-ended questions to allow participants to provide more detailed insights regarding their experiences and perceptions, both positive and negative. 

Students' prior programming experience was the only demographic question we considered relevant. Even though the course focuses on novice learners as a target group, some students do have prior programming experience due to, for example, vocational training. Finally, the survey was developed with clear and unbiased language to ensure comprehensibility and accuracy in the responses. The survey questions are available in Appendix \ref{app:survey_questions}.

\subsection{Course Context of the Data Collection}

The present study was conducted within an introductory programming course for first-year computing students at Goethe University, Frankfurt. 790 students were enrolled in that course in the winter term 2023/24. Most students were enrolled in the computer science (CS) program, even though some students pursue CS as a minor. The course is dedicated to novice learners of programming, as there are no prerequisites to participate. It is accompanied by an online course in the university's learning management system Moodle.

The class consists of a 2-hour lecture per week for all students, and a 2-hour tutorial session in groups of 20-30 students. In the tutorial, students are supposed to solve weekly or bi-weekly exercise sheets with programming problems. Even though this is voluntary, students can collect points for correct submissions (two for the exercises required for this study). These points can later be used to improve exam grades. 

For this study, a new exercise sheet was developed for one of the tutorial sessions. Students were asked to individually work on these exercises for two weeks from December 6 onwards. Group work was not allowed. Students were instructed to \say{complete the tasks using ChatGPT via the free version} (3.5) on the web interface, and to submit \say{all prompts and responses} as paired entries in a spreadsheet via the Moodle course. Students were not instructed on how to use ChatGPT-3.5. They were only provided with a link to OpenAI's guide on prompt engineering~\cite{openaiprompting}. This approach was used to avoid influencing students' use of the tool. They had only been introduced to the study's objective and the process during the lecture preceding the tutorial. 

\subsection{Task Design}

The new exercise sheet addressed several concepts, e.g., \textit{recursion}, \textit{functions}, \textit{lists}, \textit{conditionals}, \textit{string manipulation}, and \textit{documentation}. 
It comprised two tasks with sub-tasks. 

Task 1 consisted of four sub-tasks. It presented students with code snippets with recursive elements for several operations: 
\begin{itemize}
    \item[1a] summation of the digits of a number,
    \item[1b] reversing a list,
    \item[1c] performing multiplication, and
    \item[1d] computing the Ackermann function.
\end{itemize}
The task was to read and interpret the given code snippets with the goal of determining the output of the code, the number of function calls, and identifying the type of recursion.

Task 2 was designed to be more complex. It required the implementation of a function that determines the number of \say{happy strings} within all sub-strings of a given string. A \say{happy string} was defined as a string that can either be rearranged into (or already is) a repetition of some string. In particular, students were asked to:
\begin{itemize}
    \item[2a] test a given string for the \say{happy} property, and
    \item[2b] test all possible sub-strings of the given string.
\end{itemize}
Using the recursive approach was awarded with extra points.

An important criteria for the selected tasks was the performance of ChatGPT-3.5 in solving them. For example, ChatGPT demonstrated a high success rate in identifying the correct functions for the first task. However, it frequently failed in calculating the number of calls, as it often overlooked the base case. Task 2 proved to be even more challenging, and ChatGPT-3.5 never generated a correct solution in 10 regeneration attempts. It was only able to describe the necessary steps to solve the problem.

\subsection{Data Analysis}
To evaluate students' self-reported use patterns and perception of ChatGPT-3.5 in the context of solving selected programming exercises, 298 valid survey responses were analyzed. A quantitative approach was used to evaluate closed questions (Q1-Q11), while responses to open questions were qualitatively analyzed (Q12-Q14).

The survey addresses RQ1 (students' use patterns) within Q1-Q6, which are all closed questions. The quantitative analysis comprises students' programming experience prior to the introductory course in years, whether they used ChatGPT before this exercise, usage frequencies and duration, how ChatGPT was usually accessed, and for which tasks or problems it was consulted (all in the context of solving the exercise sheet). 

RQ2 on students' perceptions and user experience is addressed by survey questions Q7-Q14. Questions 7-11 refer to the ease of use of ChatGPT and other adoption criteria. The evaluation is based on the responses on the 5-point-Likert scale to indicate students' distribution and the degree to which they agree, or were satisfied. 
The remaining open questions (Q12-14) were qualitatively analyzed by applying a content analysis~\cite{mayring2000qualitative} of the responses. As each response contained multiple meaningful elements (mostly 3, as requested by Q12 and 13), each positive or negative aspect was coded once. Each student's full response to the respective open survey questions was used as context unit. 

Deductive categories reflecting students' perspective as shown in related work (see Section~\ref{sec:related_work}) constituted the starting point for building a category system (e.g., ease of use, code explanations, use as study buddy, or for debugging).
Due to the large number of responses to the open questions, and the variety of aspects, new inductive categories were built by using the psychology of text processing~\cite{ballstaedt1981texte,mandl1981psychologie}.
Thus, the responses were summarized, paraphrased and abstracted so that new categories were constructed based on the analyzed material.
Developing the inductive categories was an iterative process. After coding an initial proportion of the material (approx. 10\%), the initial categories were adapted based on students' responses. This procedure was repeated multiple times until all responses were analyzed.

\section{Results}
This section presents the students' responses to the survey questions (available online~\cite{Scholl_Kiesler_2024_datasurvey}) with the goal of answering both RQs. Students' reported use patterns of ChatGPT (RQ1) will be summarized, before introducing their perceptions of ChatGPT in the context of introductory programming exercises (RQ2). As RQ2 was addressed by closed and open questions in the survey, we outline quantitative and qualitative results.

\subsection{Students' reports on their use patterns of ChatGPT (RQ1)}

Students reported several insights into their use patterns of ChatGPT when solving the programming exercises for the introductory course. The analysis of survey questions Q1 to Q6 (n=298 each) provides a comprehensive snapshot as an answer to RQ1. 

First of all, it should be noted that the surveyed students were indeed mostly programming novices (Q1), with 34\% having no programming experience, and 43\% having limited experience of less than one year. Another 17\% of the students reported to have one to two years of programming experience before enrolling in the course, and 6\% of the students had more than 3 years of programming experience. 

When asked about students' use of ChatGPT prior to the given exercise (Q2), 84\% reported to have used ChatGPT for assignments. Only 16\% negated that question. Thus, a majority of the students in the course had little to no programming experience, yet nearly most of them had previously used ChatGPT for their coursework.
When it comes to the frequency (Q3) of using ChatGPT for their programming exercises, the data suggests that about half of the students (52\%) engage with the GenAI tool on a weekly basis, 18\% even daily. 6\% reported on using it monthly, 16\% do so rarely, and only 8\% of student have never used ChatGPT. 

Regarding duration (Q4), 43\% seem to use ChatGPT for quick requests lasting less than 15 minutes, while 38\% use it for 15-30 minutes. The remaining students use it for 30-60 minutes (16\%), or for more than 60 minutes (3\%). Students clearly prefer using the ChatGPT's interface, with 95\% of the respondents indicating this approach. Only 2\%, respectively 3\% use the messaging App integration or the API integration.  

\begin{figure}[h]
    \centering
    \includegraphics[width=0.9\linewidth]{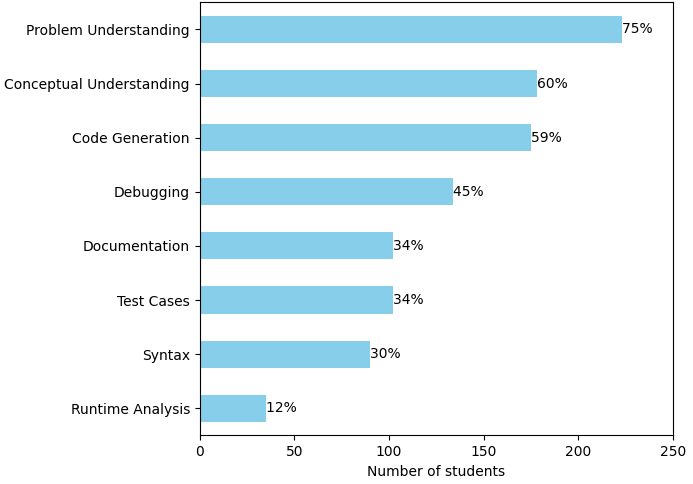}
    \caption{Responses to Q6: Tasks for which students used ChatGPT.}
    \label{fig:Q6}
\end{figure}

Students further selected all tasks for which they applied ChatGPT when working on the given programming exercises (Q6). As shown in Figure \ref{fig:Q6}, students used the tool for multiple purposes, ranging from problem understanding (223), conceptual understanding (178), code generation (176), debugging (134), producing documentation (102) and test cases (102), correcting syntax (90), to performing a runtime analysis (35). 

In addition to the closed answer options, students provided the following applications of ChatGPT: used for coding (13), to get general explanations (9), as a writing assistant (8), but also for private uses (7), as study buddy (4), search tool (4), or as a starting point when solving programming tasks (3).

\subsection{Students' perceptions of ChatGPT (RQ2)}

The second part of the survey quantitatively and qualitatively explored the subjective experiences and evaluation of the GenAI tool as reported by learners (Q7-Q14).

\subsubsection{Quantitative results}

\begin{figure*}[htb]
    \centering
    \includegraphics[width=1\textwidth]{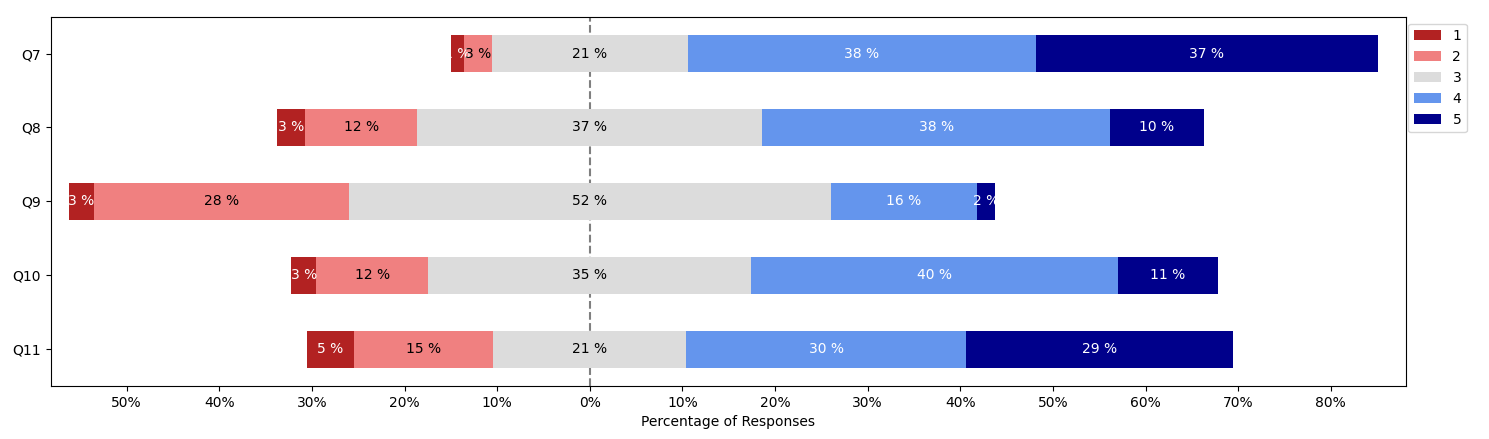}
    \caption{Likert scale assessment of student impressions on using ChatGPT (Q7-Q11). The figure displays a diverging stacked bar chart representing student ratings on ease of use, assistance in skill improvement, accuracy and relevance of responses, overall satisfaction, and likelihood to recommend ChatGPT as a learning aid. Responses range from negative (1, dark red) to positive (5, dark blue), with a neutral central point (grey).}
    \label{fig:Q7-Q11}
\end{figure*}

The diverging stacked bar chart in Figure \ref{fig:Q7-Q11} illustrates the distribution of student responses to several 5-point-Likert-scale questions (Q7-Q11, n=298 each) about their perspectives on using ChatGPT. Each row of the chart corresponds to a different question, with responses ranging from negative (left-hand side, shades of red) to positive (right-hand side, shades of blue), and a neutral midpoint (grey).

For Q7 on the ease of using ChatGPT, the median and mean of the responses were 4 and 4.06 respectively. The majority of responses were positive, with the largest proportion of students rating ChatGPT the highest on the Likert scale, indicating a perception of ChatGPT as user-friendly.

When asked about the extent to which ChatGPT has helped improve programming skills or solve coding problems (Q8), the responses were generally positive, with a median of 3 and a mean of 3.40. A significant number of students felt that ChatGPT has been helpful in these areas.

The accuracy and relevance of the responses provided by ChatGPT (Q9) yielded a more varied distribution of opinions, with a median of 3 and a mean of 2.87. These statistics suggest some students found the accuracy and relevance to be less than satisfactory, leaning towards the lower end of the scale.

Regarding overall satisfaction with ChatGPT for programming assistance (Q10), the median was 4 and the mean was 3.44, showing a positive trend. Many students expressed high levels of satisfaction with the tool.

Lastly, the likelihood of recommending ChatGPT as a support tool to a programming novice (Q11) was predominantly positive, with a median of 4 and a mean of 3.63. A substantial portion of students indicated they are very likely to recommend ChatGPT for programming novices.

Across all questions, there is a substantial number of responses in the neutral category, reflecting a degree of ambivalence or moderate views about ChatGPT among the students.

\subsubsection{Qualitative results}

The qualitative analysis of the open questions further reflect upon students' perceptions towards ChatGPT when solving programming exercises in an introductory course.
Table \ref{tab:Q12-Q13} summarizes the deductive-inductive categories of positive and negative aspects and experiences students reported in response to Q12 (n=261) and Q13 (n=259). The analysis resulted in 740 coding units with positive comments (Q12), and 682 coding units representing negative perspectives (Q13) -- 1422 in sum.

The first column of the table represents the category label, and how often students referred to it in the data, both in the positive ($\oplus$), and negative ($\ominus$) sense. The second column provides the category definition, which is followed by a literal, representative example of a student response. Overall, we identified 20 categories in the responses to Q12 and Q13.

The initial seven categories of Table \ref{tab:Q12-Q13} represent students' perspectives on several different use patterns. Among them is the use as initial help tool when starting to solve a problem, the application as a search tool for dedicated questions, or for conceptual input. Many students referred to using ChatGPT for the generation of code and text (e.g., test cases, documentation, etc.), but noted both positive and negative aspects of that scenario. Debugging is another, mostly positively mentioned use case. ChatGPT was also mentioned as producing alternative perspectives and solutions, even though students disregarded the degree of variation of the replies. Another positively mentioned use case is that of the study buddy, meaning students use ChatGPT to generate individual, on-demand assistance.

\input{tables/categories_Q12_Q13}

The remaining 13 categories presented in Table \ref{tab:Q12-Q13} summarize students' perceptions of certain qualities or aspects related to the use of ChatGPT. For example, students noted the response quality with a positive and negative connotation, e.g., due to its ability to rephrase things, but also w.r.t. its use of uncommon/unknown terms. Similarly, the chat history can help students, as they do not have to repeat themselves in a conversation. At the same time, `remembering' the conversation can require students to restart a new chat. The tool's availability is mostly mentioned in a positive manner, with downtime being criticized. The ease of using ChatGPT comprises positive aspects, such as its intuitive handling, adapting to natural language, multilingual inputs, and even vague inputs. At the same time, students are under the impression that precise inputs are required. 

ChatGPT's efficiency w.r.t. time was perceived as positive, due to the potential of saving time, and negative due to the need to verify its responses. Students further appreciated the large knowledge base, but criticized the biases it can reproduce. Another aspect concerns the perceived `social' interaction with ChatGPT. On the one hand, the safe space without `stupid' questions was appreciated. On the other hand, students noted that the tool does not comprehend emotions. 

All of the aforementioned aspects were recognized in a sort of balanced manner, expressing both benefits for students, but also criticism. However, 6 of the 13 categories only refer to negative aspects students experience using ChatGPT. Among them are privacy concerns, ChatGPT's overconfidence despite producing incorrect results, hallucinations, and a lack of integrity (e.g., by citing non-existent sources). Moreover, students expressed the necessity to remain critical of the GenAI tool, and to always verify the generated responses. Some students (26) even mentioned the risk of depending too heavily on ChatGPT, causing them to not thoroughly attempt to solve programming exercises themselves anymore.  

The last survey question (Q14) asked for additional remarks (n=136). Students mostly repeated the positive and negative aspects from Q12 and Q13. Nonetheless, we present some of the most interesting comments to highlight students' perspectives: 
\begin{myquote}
    \small
    \say{\textit{AI (and how to use it) should be introduced as a core module. If you know how to use it, you learn much faster and save a lot of energy.}} 
    \vspace{1ex}
    \\
    \say{\textit{If in the exercises the idea of how the code came up and the attempt to write it were more valued (instead of the result `the code runs well'), the process of doing the exercises might be more enjoyable.}}   
    \vspace{1ex}
    \\
    \say{\textit{I don't think it impacts my learning experience positivly. Of course it helps to get answers fast  and in my case mostly (if not all) right. But a lot of answers that stick to my head today  are the ones I have researched very long.}}
\end{myquote}

\section{Discussion}

The analysis of the responses provided by students in the survey reveal that many students incorporate ChatGPT into their routines when solving programming tasks, utilizing it for a diverse range of problems. Some of these had been identified in prior work, e.g., debugging~\cite{prather2023wgfullreport}. While many aspects of ChatGPT are perceived as helpful and useful by students, there is also notable criticism surrounding known issues such as hallucinations, overconfidence, and over-reliance.

Although some students voiced criticism regarding the occasional unavailability of ChatGPT, many valued the opportunity to receive immediate responses, along with appreciating the interactive nature of the platform. In addition, students positively highlighted the multilingual capabilities of ChatGPT and the ease of its use. They found the ability to seamlessly switch between their native language, German, and English to be particularly advantageous.

Many students expressed difficulties in understanding programming tasks and getting started with writing code. Hence, they valued the use of ChatGPT as starting point, similarly to the students surveyed in related work~\cite{vaithilingam2022expectation}. While some students indicated that ChatGPT provided satisfactory responses to straightforward prompts despite spelling or syntactic errors, a larger number of students expressed challenges in obtaining adequate responses requiring well-engineered prompts.

In terms of ChatGPT's utilization by programming novices, students voiced caution against over-reliance on these tools, which is similar to the findings in prior work~\cite{xue2024does,barke2023grounded}. Conversely, other students recognize increasing expectations by programming educators and see ChatGPT as a means to counteract this. Additionally, there is acknowledgment of other models, like ChatGPT-4 or Copilot, which may offer enhanced features, e.g., improved effectiveness or support for image/file input. However, some students are concerned about the associated costs, which potentially increases the disparities in the equal access and affordability of education. So far, GenAI does not seem to have democratized education.

\section{Limitations}

The students' responses proved sufficient to reveal their use patterns and perspectives on GenAI tools, such as ChatGPT. However, the limits of self-reporting (e.g., social desirability, exaggerations, omissions) apply to the present work. Moreover, as students knew they were surveyed, the observer's paradox~\cite{roethlisberger1939management} needs to be taken into account when interpreting the data. Another limitation refers to the course setting, the selected tasks, and the institution where the data was gathered (i.e., in only one country). Expanding or replicating this study in other areas is considered future work.

\section{Conclusions and Future Work}

The present research pursued the goal of gaining insights into the student perspective of using GenAI tools, such as ChatGPT, in the context of a curricular introductory programming course. To identify students' use patterns along with their perceptions of the tool, we designed an exercise sheet with programming tasks for novice learners at a German higher education institution. Students were asked to use ChatGPT-3.5 (freely available at the time, in Dec. 2023) without further instruction. After submitting their solutions to the programming exercises, students (n=298) filled out an online survey to reflect upon their use and impressions. 

The results revealed that roughly half of the students use ChatGPT as often as every week, 18\% even use it daily. We further identified a great diversity of students' use patterns and application scenarios, which is due to the sheer capacity of GenAI. For example, students use it to get started with their assignments, to generate text and code, for debugging, or as a search tool when having questions. At the same time, students' perceptions showed their ambivalence towards GenAI, as students seem to be well aware of its limitations and the potential (negative) impact on student learning, and equal opportunities. It is therefore crucial for educators to at least acknowledge GenAI tools in computing and programming education, and to provide guardrails and instructions for students on how to navigate through the GenAI revolution. 

There are several pathways for future work. Among them is the triangulation of the present work with chat protocols from novice learners of programming with a GenAI tool to address the limitations of self-reporting~\cite{scholl2024analyzing}. Another avenue is the development of specialized GPT models incorporated into well-known educational environments, to offer students free, straightforward access. Such models could contain built-in guardrails to facilitate a guided use. Integrating prompt-recommender-systems could also assist students in optimizing their queries. In the long run, it is essential to not only offer tools as resource but to integrate them into educational setting, evaluate them, and reflect upon their use.

\appendix{}
\subsection{Survey Questions}
\label{app:survey_questions}
\footnotesize
\begin{enumerate}
    \item[Q1] Programming Experience before Course.
    \item[Q2] Did you use ChatGPT prior to this exercise for assignments?
    \item[Q3] How often do you use ChatGPT?
    \item[Q4] On average, how long do you engage with ChatGPT in a single session?
    \item[Q5] Which platform do you primarily use for accessing ChatGPT?
    \item[Q6] Please select all the tasks for which you used ChatGPT. (Check all that apply)
    \item[Q7] How would you rate the ease of using ChatGPT? (Likert scale, 1 Very difficult - 5 Very easy)
    \item[Q8] To what extent has ChatGPT helped in improving your programming skills or solving coding problems? (Likert scale, 1 Not at all - 5 Greatly improved)
    \item[Q9] Rate the accuracy and relevance of the responses provided by ChatGPT. (Likert scale, 1 Very inaccurate - 5 Very accurate)
    \item[Q10] How satisfied are you with your overall experience using ChatGPT for programming assistance? (Likert scale, 1 Not at all - Very satisfied)
    \item[Q11] How likely are you to recommend ChatGPT as a support tool to a programming novice? (Likert scale, 1 Not at all - 5 Highly likely)
    \item[Q12] Please share three positive aspects or examples of your experience using ChatGPT. What did you find most valuable or beneficial? (open)
    \item[Q13] Please share three negative aspects or examples of your experience using ChatGPT. What did you find challenging or difficult? (open)
    \item[Q14] Is there anything else you would like to share about your experience using ChatGPT? (open)
\end{enumerate}

\bibliographystyle{IEEEtran}
\bibliography{literature}

\end{document}

%% file: tables/categories_Q12_Q13.tex
\begin{table*}[t!]
\centering \scriptsize
\caption{Categories, definitions, and anchor examples reflecting students' perceptions (RQ2).}
\label{tab:Q12-Q13}
\resizebox{\linewidth}{!}{ 
\begin{tabular}{@{}lll@{}}
\toprule
\textbf{Category} & \textbf{Definition} & \textbf{Anchor Example(s)} \\ \midrule 
    \makecell[l]{Starting Point \\ $\oplus$45} 
    & \makecell[l]{Refers to initial guidance and structure provided by ChatGPT,\\ including code templates.} 
    & \makecell[l]{$\oplus$\say{Provides at least an approach when one has no idea where to begin.}}\\ \hline

    \makecell[l]{Conceptual Input \\ $\oplus$59} 
    & \makecell[l]{Refers to ChatGPT's ability to provide new concepts, ideas and\\inspiration, including suggestions for functions, libraries and hints.} 
    & \makecell[l]{$\oplus$\say{If one suspects that there is a certain command or built-in function\\ that most likely exists but does not know it.}} \\ \hline

    \makecell[l]{Study Buddy \\ $\oplus$23} 
    & \makecell[l]{Refers to ChatGPT as companion-like support tool providing\\ personalized guidance, tutoring, and interactive assistance.} 
    & \makecell[l]{$\oplus$\say{Offers instant help with coding problems and advice on best practices,\\ acting like an on-demand mentor.}} \\ \hline

    \makecell[l]{Code and Text\\ Generation \\ $\oplus$89 \quad $\ominus$111} 
    & \makecell[l]{Refers to ChatGPT's ability to produce, transform, or finalize\\ various forms of content, such as code, text, tests or documentation.} 
    & \makecell[l]{$\oplus$\say{I can work well with the code that ChatGPT provides.}\\
                   $\ominus$\say{More complex code is almost unusable.}} \\ \hline

    \makecell[l]{Debugging \\ $\oplus$85 \quad $\ominus$5} 
    & \makecell[l]{Refers to assisting with troubleshooting and refining code input by\\ identifying and addressing compile, runtime, syntax and logical errors.} 
    & \makecell[l]{$\oplus$\say{ChatGPT is very good at identifying minor errors in the code 
    }\\
                   $\ominus$\say{When reviewing, ChatGPT itself creates mistakes.}} \\ \hline

    \makecell[l]{Alternative \\Perspectives\\ and Solutions \\ $\oplus$18\ \quad $\ominus$18} 
    & \makecell[l]{Refers to ChatGPT's capacity to offer students a range of opinions,\\ perspectives, and approaches to programming tasks; includes providing\\ alternative solutions and varied rephrasing of questions or answers.} 
    & \makecell[l]{$\oplus$\say{A different perspective on the task (different code).}\\
                   $\ominus$\say{Often provides incorrect answers or different answers to the same\\ question upon repeated input.}} \\ \hline

    \makecell[l]{Search Tool \\ $\oplus$29\  \quad $\ominus$1} 
    & \makecell[l]{Refers to use of ChatGPT as a resource for gathering information\\ and finding answers, similar to search engines and online forums.} 
    & \makecell[l]{$\oplus$\say{It shortens my time spent searching for information on the internet\\and helps me avoid wasting time looking for unnecessary information.}\\
                   $\ominus$\say{The outputs often do not go beyond Google search results.}} \\ \hline \hline

    \makecell[l]{Response Quality \\ $\oplus$146 \quad  \hspace{-0.85em} $\ominus$142} 
    & \makecell[l]{Refers to the level of accuracy, clarity, comprehensibility and\\ usefulness as well as structure and format of ChatGPT's answers to\\ students' questions.} 
    & \makecell[l]{$\oplus$\say{
    If a task is not entirely clear, ChatGPT can rephrase it differently.}\\
                   $\ominus$\say{Sometimes the content is not explained well, and strange\\ terms and/or explanations are used.}} \\ \hline

    \makecell[l]{Chat History \\ $\oplus$27\ \quad $\ominus$79} 
    & \makecell[l]{Refers to ChatGPT's ability to retain previous interactions,\\ enabling users to revisit and correct prior queries and responses.} 
    & \makecell[l]{$\oplus$\say{The program learns during the conversation [...] 
    the context does not\\ need to be re-explained with each question.}\\
                   $\ominus$\say{ChatGPT often needs to be restarted because it remembers `answers'.}}\\ \hline 
    \makecell[l]{Availability \\ $\oplus$84 \quad $\ominus$35} 
    & \makecell[l]{Refers to prerequisites required to access ChatGPT (i.e., account/\\subscription) available servers, and to ChatGPT's response times.} 
    & \makecell[l]{$\oplus$\say{It is quick [...] A large amount of code can be generated. It is free.}\\
                   $\ominus$\say{ChatGPT is often down and unusable.}} \\ \hline

    \makecell[l]{Ease of Use \\ $\oplus$67 \quad $\ominus$119} 
    & \makecell[l]{Refers to the level of accessibility and user-friendliness; includes\\ the ability to adapt to natural language inputs, and the structure,\\ language and precision of instructions.} 
    & \makecell[l]{$\oplus$\say{Even vaguely stated questions were explained thoroughly and precisely.}\\ 
                   $\ominus$\say{The AI often fails when it is not provided with enough data. Precise\\ formulation is necessary.}} \\ \hline

    \makecell[l]{Time Efficiency \\ $\oplus$29 \quad $\ominus$10} 
    & \makecell[l]{Refers to the potential of ChatGPT to save time by providing quick\\ answers and reducing workload, despite having to verify responses.} 
    & \makecell[l]{$\oplus$\say{Routine tasks are completed quickly.}\\
                   $\ominus$\say{Slower for more complex problems than writing the code oneself.}} \\ \hline

    \makecell[l]{Knowledge Base \\ $\oplus$31 \quad $\ominus$29} 
    & \makecell[l]{Refers to range and scope of available information based on training\\ data; includes versatility, relevance and to-dateness of the information.} 
    & \makecell[l]{$\oplus$\say{ChatGPT is based on a large training dataset.}\\
                   $\ominus$\say{It is biased on non-purely technical topics.}} \\ \hline

    \makecell[l]{Social Interaction \\ $\oplus$8 \hspace{0.1em} \quad $\ominus$9} 
    & \makecell[l]{Refers to the manner in which ChatGPT engages with students,\\ focusing on the conversational aspects and the overall experience\\ of communication.} 
    & \makecell[l]{$\oplus$\say{With ChatGPT, there are no stupid questions; it is a safe space.}\\
                   $\ominus$\say{The model may not understand emotions well and could provide\\ inappropriate responses in emotional situations.}} \\ \hline

    \makecell[l]{Privacy Concerns \\ \quad \hspace{1.8em} $\ominus$5} 
    & \makecell[l]{Refers to the assumptions students may have about the\\ confidentiality/privacy of their interactions with ChatGPT.} 
    & \makecell[l]{$\ominus$\say{I am constantly concerned that my messages are read by third parties\\ and always feel an underlying unease when using it.}} \\ \hline

    \makecell[l]{Overconfidence \\  \quad \hspace{1.8em} $\ominus$29} 
    & \makecell[l]{Refers to ChatGPT's presenting information with a high level\\ of certainty, even when the data may be incorrect or questionable.} 
    & \makecell[l]{$\ominus$\say{Questions can be just outright wrong, with the language model\\ insisting it was correct.}} \\ \hline

    \makecell[l]{Hallucinations \\ \quad \hspace{1.8em} $\ominus$16} 
    & \makecell[l]{Refers to ChatGPT generating responses that include inaccurate\\ or fabricated information, such as citing non-existent sources.} 
    & \makecell[l]{$\ominus$\say{I once asked for literature recommendations. The responses seemed\\ credible, but they were entirely fabricated.}} \\ \hline

    \makecell[l]{Lack of Integrity\\ \quad \hspace{1.8em} $\ominus$12} 
    & \makecell[l]{Refers to clarity and reliability of the provided information,\\ including references to credible sources ensuring academic integrity.} 
    & \makecell[l]{$\ominus$\say{Unverifiable or no information about sources.}} \\ \hline

    \makecell[l]{Criticizing GenAI \\ \quad \hspace{1.8em} $\ominus$36} 
    & \makecell[l]{Refers to students having to critically evaluate and verify ChatGPT's\\ responses.} 
    & \makecell[l]{$\ominus$\say{One must be able to judge whether the solution is correct or not. ChatGPT\\ should only be used if one already has basic programming knowledge.}} \\ \hline

    \makecell[l]{GenAI dependency\\ \quad \hspace{1.8em} $\ominus$26} 
    & \makecell[l]{Refers to risk for students to depend too heavily on ChatGPT\\ for programming exercises.} 
    & \makecell[l]{$\ominus$\say{When I have a problem, I don't think about it for as long as I used to,\\ I just go and ask a ChatGPT. It made me depend on it.}} \\

\bottomrule
\end{tabular}
} 
\end{table*}

%% file: chapgpt_fie_2023.bbl
\begin{thebibliography}{10}
\providecommand{\url}[1]{#1}
\csname url@samestyle\endcsname
\providecommand{\newblock}{\relax}
\providecommand{\bibinfo}[2]{#2}
\providecommand{\BIBentrySTDinterwordspacing}{\spaceskip=0pt\relax}
\providecommand{\BIBentryALTinterwordstretchfactor}{4}
\providecommand{\BIBentryALTinterwordspacing}{\spaceskip=\fontdimen2\font plus
\BIBentryALTinterwordstretchfactor\fontdimen3\font minus \fontdimen4\font\relax}
\providecommand{\BIBforeignlanguage}[2]{{%
\expandafter\ifx\csname l@#1\endcsname\relax
\typeout{** WARNING: IEEEtran.bst: No hyphenation pattern has been}%
\typeout{** loaded for the language `#1'. Using the pattern for}%
\typeout{** the default language instead.}%
\else
\language=\csname l@#1\endcsname
\fi
#2}}
\providecommand{\BIBdecl}{\relax}
\BIBdecl

\bibitem{amani2023generative}
S.~Amani, L.~White, T.~Balart, L.~Arora, D.~K.~J. Shryock, D.~K. Brumbelow, and D.~K.~L. Watson, ``{Generative AI Perceptions: A Survey to Measure the Perceptions of Faculty, Staff, and Students on Generative AI Tools in Academia},'' 2023.

\bibitem{raman2023university}
R.~Raman, S.~Mandal, P.~Das, T.~Kaur, J.~Sanjanasri, and P.~Nedungadi, ``{University students as early adopters of ChatGPT: Innovation Diffusion Study},'' 2023.

\bibitem{prather2023wgfullreport}
J.~Prather, P.~Denny, J.~Leinonen, B.~A. Becker, I.~Albluwi, M.~Craig, H.~Keuning, N.~Kiesler, T.~Kohn, A.~Luxton-Reilly, S.~MacNeil, A.~Petersen, R.~Pettit, B.~N. Reeves, and J.~Savelka, ``{The Robots Are Here: Navigating the Generative AI Revolution in Computing Education},'' in \emph{Proceedings of the 2023 Working Group Reports on Innovation and Technology in Computer Science Education}, ser. ITiCSE-WGR '23.\hskip 1em plus 0.5em minus 0.4em\relax New York: ACM, 2023, p. 108–159.

\bibitem{kiesler2024modeling}
\BIBentryALTinterwordspacing
N.~Kiesler, \emph{Modeling Programming Competency: A Qualitative Analysis}.\hskip 1em plus 0.5em minus 0.4em\relax Cham: Springer International Publishing, 2024. [Online]. Available: \url{https://doi.org/10.1007/978-3-031-47148-3}
\BIBentrySTDinterwordspacing

\bibitem{becker2023generative}
\BIBentryALTinterwordspacing
B.~A. Becker, M.~Craig, P.~Denny, H.~Keuning, N.~Kiesler, J.~Leinonen, A.~Luxton-Reilly, J.~Prather, and K.~Quille, ``{Generative AI in Introductory Programming},'' 2023. [Online]. Available: \url{https://csed.acm.org/wp-content/uploads/2023/12/Generative-AI-Nov-2023-Version.pdf}
\BIBentrySTDinterwordspacing

\bibitem{kiesler2023beyond}
N.~Kiesler, ``{Beyond the Textbook: Rethinking Students' Competencies in the LLM Era},'' Generative AI: Implications for Teaching and Learning, Uppsala, Sweden, 2023.

\bibitem{geng2023chatgpt}
C.~Geng, Y.~Zhang, B.~Pientka, and X.~Si, ``{Can ChatGPT Pass An Introductory Level Functional Language Programming Course?}'' 2023.

\bibitem{kiesler2023large}
N.~Kiesler and D.~Schiffner, ``{Large Language Models in Introductory Programming Education: ChatGPT's Performance and Implications for Assessments},'' 2023, 10.48550/arXiv.2308.08572.

\bibitem{savelka2023large}
J.~Savelka, A.~Agarwal, C.~Bogart, and M.~Sakr, ``{Large Language Models (GPT) Struggle to Answer Multiple-Choice Questions about Code},'' 2023.

\bibitem{Leinonen2023}
J.~Leinonen, A.~Hellas, S.~Sarsa, B.~Reeves, P.~Denny, J.~Prather, and B.~A. Becker, ``{Using Large Language Models to Enhance Programming Error Messages},'' in \emph{Proc. SIGCSE}.\hskip 1em plus 0.5em minus 0.4em\relax ACM, Mar. 2023.

\bibitem{Sarsa2022}
S.~Sarsa, P.~Denny, A.~Hellas, and J.~Leinonen, ``{Automatic Generation of Programming Exercises and Code Explanations Using Large Language Models},'' in \emph{Proc. ICER}.\hskip 1em plus 0.5em minus 0.4em\relax ACM, Aug. 2022.

\bibitem{macneil2022experiences}
S.~MacNeil, A.~Tran, A.~Hellas, J.~Kim, S.~Sarsa, P.~Denny, S.~Bernstein, and J.~Leinonen, ``{Experiences from Using Code Explanations Generated by Large Language Models in a Web Software Development E-Book},'' in \emph{Proc. SIGCSE TS}, 2023, p. 931–937.

\bibitem{leinonen2023comparing}
J.~Leinonen, P.~Denny, S.~MacNeil, S.~Sarsa, S.~Bernstein, J.~Kim, A.~Tran, and A.~Hellas, ``{Comparing Code Explanations Created by Students and Large Language Models},'' in \emph{Proc. ITiCSE}, 2023, pp. 124--130.

\bibitem{Bengtsson_Kaliff_2023}
\BIBentryALTinterwordspacing
D.~Bengtsson and A.~Kaliff, ``{Assessment Accuracy of a Large Language Model on Programming Assignments},'' 2023. [Online]. Available: \url{https://urn.kb.se/resolve?urn=urn:nbn:se:kth:diva-331000}
\BIBentrySTDinterwordspacing

\bibitem{kiesler2023exploring}
N.~Kiesler, D.~Lohr, and H.~Keuning, ``{Exploring the Potential of Large Language Models to Generate Formative Programming Feedback},'' in \emph{2023 IEEE Frontiers in Education Conference (FIE)}, 2024, pp. 1--5.

\bibitem{LMU-TEL/ADS2023}
I.~Azaiz, O.~Deckarm, and S.~Strickroth, ``{AI-enhanced Auto-Correction of Programming Exercises: How Effective is GPT-3.5?}'' \emph{International Journal of Engineering Pedagogy (iJEP)}, vol.~13, no.~8, pp. 67--83, Dec. 2023.

\bibitem{azaiz2024feedbackgeneration}
I.~Azaiz, N.~Kiesler, and S.~Strickroth, ``{Feedback-Generation for Programming Exercises With GPT-4},'' 2024.

\bibitem{roest2023nextstep}
L.~Roest, H.~Keuning, and J.~Jeuring, ``{Next-Step Hint Generation for Introductory Programming Using Large Language Models},'' 2023.

\bibitem{jeuring2022towards}
J.~Jeuring, H.~Keuning, S.~Marwan, D.~Bouvier, C.~Izu, N.~Kiesler, T.~Lehtinen, D.~Lohr, A.~Peterson, and S.~Sarsa, ``{Towards Giving Timely Formative Feedback and Hints to Novice Programmers},'' in \emph{Proceedings of the 2022 Working Group Reports on Innovation and Technology in Computer Science Education}.\hskip 1em plus 0.5em minus 0.4em\relax New York: ACM, 2022, p. 95–115.

\bibitem{kiesler2023investigating}
N.~Kiesler, ``{Investigating the Use and Effects of Feedback in CodingBat Exercises: An Exploratory Thinking Aloud Study},'' in \emph{2023 Future of Educational Innovation-Workshop Series Data in Action}, 2023, pp. 1--12.

\bibitem{alhossami2024socratic}
E.~Al-Hossami, R.~Bunescu, J.~Smith, and R.~Teehan, ``{Can Language Models Employ the Socratic Method? Experiments with Code Debugging},'' in \emph{Proceedings of the 55th ACM Technical Symposium on Computer Science Education V. 1}.\hskip 1em plus 0.5em minus 0.4em\relax New York: ACM, 2024, p. 53–59.

\bibitem{joshi2024chatgpt}
I.~Joshi, R.~Budhiraja, H.~Dev, J.~Kadia, M.~O. Ataullah, S.~Mitra, H.~D. Akolekar, and D.~Kumar, ``{ChatGPT in the Classroom: An Analysis of Its Strengths and Weaknesses for Solving Undergraduate Computer Science Questions},'' in \emph{Proceedings of the 55th ACM Technical Symposium on Computer Science Education V. 1}, ser. SIGCSE 2024.\hskip 1em plus 0.5em minus 0.4em\relax New York: ACM, 2024, p. 625–631.

\bibitem{prather2024howinstructors}
J.~Prather, J.~Leinonen, N.~Kiesler, J.~G. Benario, S.~Lau, S.~MacNeil, N.~Norouzi, S.~Opel, V.~Pettit, L.~Porter, B.~N. Reeves, J.~Savelka, D.~H. Smith, S.~Strickroth, and D.~Zingaro, ``{How Instructors Incorporate Generative AI into Teaching Computing},'' in \emph{Proceedings of the 2024 on Innovation and Technology in Computer Science Education V. 2}, ser. ITiCSE 2024.\hskip 1em plus 0.5em minus 0.4em\relax New York: ACM, 2024, p. 771–772.

\bibitem{amoozadeh2024trust}
M.~Amoozadeh, D.~Daniels, D.~Nam, A.~Kumar, S.~Chen, M.~Hilton, S.~Srinivasa~Ragavan, and M.~A. Alipour, ``{Trust in Generative AI among Students: An exploratory study},'' in \emph{Proceedings of the 55th ACM Technical Symposium on Computer Science Education V. 1}, ser. SIGCSE 2024.\hskip 1em plus 0.5em minus 0.4em\relax New York: ACM, 2024, p. 67–73.

\bibitem{rogers2024attitudes}
M.~P. Rogers, H.~M. Hillberg, and C.~L. Groves, ``{Attitudes Towards the Use (and Misuse) of ChatGPT: A Preliminary Study},'' in \emph{Proceedings of the 55th ACM Technical Symposium on Computer Science Education V. 1}, ser. SIGCSE 2024.\hskip 1em plus 0.5em minus 0.4em\relax New York: ACM, 2024, p. 1147–1153.

\bibitem{liu2024teaching}
R.~Liu, C.~Zenke, C.~Liu, A.~Holmes, P.~Thornton, and D.~J. Malan, ``{Teaching CS50 with AI: Leveraging Generative Artificial Intelligence in Computer Science Education},'' in \emph{Proc. SIGCSE}.\hskip 1em plus 0.5em minus 0.4em\relax New York: ACM, 2024, p. 750–756.

\bibitem{grande2024studentperspective}
V.~Grande, N.~Kiesler, and M.~A.~F. Rodriguez, ``{Student Perspectives on Using a Large Language Model (LLM) for an Assignment on Professional Ethics},'' in \emph{Proceedings of the 2024 Conference on Innovation and Technology in Computer Science Education V. 2}, ser. ITiCSE 2024.\hskip 1em plus 0.5em minus 0.4em\relax New York: ACM, 2024.

\bibitem{macneil2024discussing}
S.~MacNeil, J.~Leinonen, P.~Denny, N.~Kiesler, A.~Hellas, J.~Prather, B.~A. Becker, M.~Wermelinger, and K.~Reid, ``{Discussing the Changing Landscape of Generative AI in Computing Education},'' in \emph{Proceedings of the 55th ACM Technical Symposium on Computer Science Education V. 2}, ser. SIGCSE 2024.\hskip 1em plus 0.5em minus 0.4em\relax New York: ACM, 2024, p. 1916.

\bibitem{finnieansley22}
J.~Finnie-Ansley, P.~Denny, B.~A. Becker, A.~Luxton-Reilly, and J.~Prather, ``{The Robots Are Coming: Exploring the Implications of OpenAI Codex on Introductory Programming},'' in \emph{Proc. ACE}.\hskip 1em plus 0.5em minus 0.4em\relax New York: ACM, 2022, p. 10–19.

\bibitem{finnieansley23}
J.~Finnie-Ansley, P.~Denny, A.~Luxton-Reilly, E.~A. Santos, J.~Prather, and B.~A. Becker, ``{My AI Wants to Know if This Will Be on the Exam: Testing OpenAI’s Codex on CS2 Programming Exercises},'' in \emph{Proc. ACE}.\hskip 1em plus 0.5em minus 0.4em\relax New York: ACM, 2023, p. 97–104.

\bibitem{parlante2024}
\BIBentryALTinterwordspacing
N.~Parlante, ``{CodingBat},'' Online Publication, 2024. [Online]. Available: \url{https://codingbat.com/about.html}
\BIBentrySTDinterwordspacing

\bibitem{gill2024transformative}
S.~S. Gill, M.~Xu, P.~Patros, H.~Wu, R.~Kaur, K.~Kaur, S.~Fuller, M.~Singh, P.~Arora, A.~K. Parlikad, V.~Stankovski, A.~Abraham, S.~K. Ghosh, H.~Lutfiyya, S.~S. Kanhere, R.~Bahsoon, O.~Rana, S.~Dustdar, R.~Sakellariou, S.~Uhlig, and R.~Buyya, ``{Transformative effects of ChatGPT on modern education: Emerging Era of AI Chatbots},'' \emph{Internet of Things and Cyber-Physical Systems}, vol.~4, pp. 19--23, 2024.

\bibitem{zhai2022chatgpt}
X.~Zhai, ``{ChatGPT User Experience: Implications for Education},'' 2022.

\bibitem{zastudil2023studentpersp}
C.~Zastudil, M.~Rogalska, C.~Kapp, J.~Vaughn, and S.~MacNeil, ``{Generative AI in Computing Education: Perspectives of Students and Instructors},'' in \emph{2023 IEEE Frontiers in Education Conference}, 2023, pp. 1--9.

\bibitem{phung2023generating}
T.~Phung, J.~Cambronero, S.~Gulwani, T.~Kohn, R.~Majumdar, A.~Singla, and G.~Soares, ``{Generating High-Precision Feedback for Programming Syntax Errors using Large Language Models},'' 2023.

\bibitem{zhang2022repairing}
J.~Zhang, J.~Cambronero, S.~Gulwani, V.~Le, R.~Piskac, G.~Soares, and G.~Verbruggen, ``{Repairing Bugs in Python Assignments Using Large Language Models},'' \emph{arXiv preprint arXiv:2209.14876}, 2022.

\bibitem{narciss2006}
S.~Narciss, \emph{{Informatives Tutorielles Feedback: Entwicklungs- und Evaluationsprinzipien auf der Basis instruktionspsychologischer Erkenntnisse}}.\hskip 1em plus 0.5em minus 0.4em\relax M\"{u}nster: Waxmann Verlag, 2006.

\bibitem{taylor2024dcchelperrorexplanations}
A.~Taylor, A.~Vassar, J.~Renzella, and H.~Pearce, ``{dcc --help: Transforming the Role of the Compiler by Generating Context-Aware Error Explanations with Large Language Models},'' in \emph{Proc. SIGCSE}.\hskip 1em plus 0.5em minus 0.4em\relax New York: ACM, 2024, p. 1314–1320.

\bibitem{kazemitabaar2024codeaid}
M.~Kazemitabaar, R.~Ye, X.~Wang, A.~Z. Henley, P.~Denny, M.~Craig, and T.~Grossman, ``{CodeAid: Evaluating a Classroom Deployment of an LLM-based Programming Assistant that Balances Student and Educator Needs},'' \emph{arXiv preprint arXiv:2401.11314}, 2024.

\bibitem{prather2023s}
J.~Prather, B.~N. Reeves, P.~Denny, B.~A. Becker, J.~Leinonen, A.~Luxton-Reilly, G.~Powell, J.~Finnie-Ansley, and E.~A. Santos, ``{“It’s Weird That It Knows What I Want”: Usability and Interactions with Copilot for Novice Programmers},'' \emph{ACM Trans. Comput.-Hum. Interact.}, 2023.

\bibitem{vaithilingam2022expectation}
P.~Vaithilingam, T.~Zhang, and E.~L. Glassman, ``{Expectation vs. Experience: Evaluating the Usability of Code Generation Tools Powered by Large Language Models}.''\hskip 1em plus 0.5em minus 0.4em\relax New York: ACM, 2022, pp. 1--7.

\bibitem{jayagopal2022exploring}
D.~Jayagopal, J.~Lubin, and S.~E. Chasins, ``{Exploring the learnability of program synthesizers by novice programmers},'' in \emph{Proc. ACM Symposium on User Interface Software and Technology}, 2022, pp. 1--15.

\bibitem{denny2023promptly}
P.~Denny, J.~Leinonen, J.~Prather, A.~Luxton-Reilly, T.~Amarouche, B.~A. Becker, and B.~N. Reeves, ``{Promptly: Using Prompt Problems to Teach Learners How to Effectively Utilize AI Code Generators},'' 2023.

\bibitem{barke2023grounded}
S.~Barke, M.~B. James, and N.~Polikarpova, ``{Grounded Copilot: How Programmers Interact with Code-Generating Models},'' \emph{Proc. ACM Program. Lang.}, vol.~7, no. OOPSLA1, 4 2023.

\bibitem{xue2024does}
\BIBentryALTinterwordspacing
Y.~Xue, H.~Chen, G.~R. Bai, R.~Tairas, and Y.~Huang, ``{Does ChatGPT Help With Introductory Programming? An Experiment of Students Using ChatGPT in CS1},'' 2024. [Online]. Available: \url{https://yuhuang-lab.github.io/paper/ICSE-SEET2024.pdf}
\BIBentrySTDinterwordspacing

\bibitem{Forman_Udvaros_Avornicului_2023}
N.~Forman, J.~Udvaros, and M.~S. Avornicului, ``{ChatGPT: A new study tool shaping the future for high school students},'' \emph{International Journal of Advanced Natural Sciences and Engineering Researches}, vol.~7, no.~4, p. 95–102, May 2023.

\bibitem{openaiprompting}
OpenAI, ``{Prompt Engineering},'' \url{https://platform.openai.com/docs/guides/prompt-engineering}, 2023, accessed: 2024-04-08.

\bibitem{mayring2000qualitative}
P.~Mayring, ``Qualitative content analysis,'' \emph{{Forum: qualitative social research}}, vol.~1, no.~2, 2000.

\bibitem{ballstaedt1981texte}
S.-P. Ballstaedt, H.~Mandl, W.~Schnotz, and S.-O. Tergan, \emph{{Texte verstehen, Texte gestalten}}.\hskip 1em plus 0.5em minus 0.4em\relax M{\"u}nchen: Urban u. Schwarzenberg, 1981.

\bibitem{mandl1981psychologie}
H.~Mandl, \emph{{Zur Psychologie der Textverarbeitung: Ans{\"a}tze, Befunde, Probleme}}.\hskip 1em plus 0.5em minus 0.4em\relax M{\"u}nchen: Urban \& Schwarzenberg, 1981.

\bibitem{Scholl_Kiesler_2024_datasurvey}
\BIBentryALTinterwordspacing
A.~Scholl and N.~Kiesler, ``{Data: Analyzing Chat Protocols of Novice Programmers Solving Introductory Programming Tasks with ChatGPT},'' Jul 2024. [Online]. Available: \url{osf.io/wbkqv}
\BIBentrySTDinterwordspacing

\bibitem{roethlisberger1939management}
F.~J. Roethlisberger and W.~J. Dickson, \emph{{Management and the Worker}}.\hskip 1em plus 0.5em minus 0.4em\relax Cambridge: Harvard University Press, 1939.

\bibitem{scholl2024analyzing}
A.~Scholl, D.~Schiffner, and N.~Kiesler, ``{Analyzing Chat Protocols of Novice Programmers Solving Introductory Programming Tasks with ChatGPT},'' 2024, 10.48550/arXiv.2405.19132.

\end{thebibliography}
